\documentclass[review]{elsarticle}

\usepackage{lineno,hyperref}
\usepackage{bm}
\usepackage{amsmath}
\usepackage{xcolor}
\usepackage{enumitem}
\usepackage{comment}
\usepackage{subfig}
\modulolinenumbers[5]
\usepackage[ruled,vlined]{algorithm2e}
\include{pythonlisting}

\journal{Journal of \LaTeX\ Templates}

\begin{document}
\begin{frontmatter}
\title{When and Where to Step: Terrain-Aware Real-Time Footstep Location and Timing Optimization for Bipedal Robots\tnoteref{mytitlenote}}

\author[firstaddress]{Ke Wang\corref{mycorrespondingauthor}}
\author[firstaddress,secondaddress]{Zhaoyang Jacopo Hu\corref{equalcontributon}}
\author[firstaddress,secondaddress]{Peter Tisnikar\corref{equalcontributon}}
\author[thirdaddress]{Oskar Helander\corref{equalcontributon}}
\author[firstaddress]{Digby Chappell}
\author[firstaddress]{Petar Kormushev}


\cortext[mycorrespondingauthor]{Corresponding author}
\cortext[equalcontributon]{Authors have contributed equally to this work}
\ead{k.wang17@imperial.ac.uk}

\address[firstaddress]{Robot Intelligence Lab, Dyson School of Design Engineering, Imperial College London}
\address[secondaddress]{Department of Bioengineering, Imperial College London}
\address[thirdaddress]{Vision for Robotics Lab, Department of Mechanical and Process Engineering, ETH Z\"{u}rich}
\begin{abstract}
Online footstep planning is essential for bipedal walking robots, allowing them to walk in the presence of disturbances and sensory noise. Most of the literature on the topic has focused on optimizing the footstep placement while keeping the step timing constant. In this work, we introduce a footstep planner capable of optimizing footstep placement and step time online. The proposed planner, consisting of an Interior Point Optimizer (IPOPT) and an optimizer based on Augmented Lagrangian (AL) method with analytical gradient descent, solves the full dynamics of the Linear Inverted Pendulum (LIP) model in real time to optimize for footstep location as well as step timing at the rate of 200~Hz. We show that such asynchronous real-time optimization with the AL method (ARTO-AL) provides the required robustness and speed for successful online footstep planning. Furthermore, ARTO-AL can be extended to plan footsteps in 3D, allowing terrain-aware footstep planning on uneven terrains. Compared to an algorithm with no footstep time adaptation, our proposed ARTO-AL demonstrates increased stability in simulated walking experiments as it can resist pushes on flat ground and on a $10^{\circ}$ ramp up to 120 N and 100 N respectively. Videos\protect\footnote{https://youtu.be/ABdnvPqCUu4} and open-source code\protect\footnote{https://github.com/WangKeAlchemist/ARTO-AL/tree/master} are released.

\end{abstract}


\end{frontmatter}


\section{INTRODUCTION}

In order for legged robots to dynamically balance themselves in the presence of disturbance, it is essential for them to plan footstep positioning and duration in real time. Much of existing work focuses on optimizing footstep locations while keeping step timing fixed, according to the dynamics of LIP model \cite{Motoi2009, Kajita2010}. The LIP model is a widely used reduced-order model of the dynamics of a bipedal walking robot; the legs are assumed to be massless, and the body is a concentrated point mass located at the robot's centre of mass (CoM) and the height of the CoM with respect to the ground is assumed to be constant \cite{Kajita2001}. Using this model, online footstep planning can be characterised by a real-time nonlinear Model Predictive Control (MPC) problem.
Due to the second order differential equation governing the LIP model, optimizing footstep duration with respect to time produces a non-linear optimization problem which is difficult to solve in real time. Thus, real-time footstep timing adaptation work has largely been focused on simplifying the dynamics used in optimization, or carefully tailoring heuristic solutions.

One popular method to simplify the control problem is to consider only the Divergent Component of Motion (DCM) of the body and predict only 'one step' ahead. By only considering the DCM of the next step, dynamics are reduced to a first order problem, where the optimal step time to control the DCM about a nominal point can be computed with Quadratic Programming (QP) \cite{Khadiv2016, Khadiv2020}. DCM can also be combined with the virtual repellent point \cite{Englsberger2013} to encode all forces on a robot's body in one single point. \cite{Englsberger2017} controls the DCM in this way by interpolating the virtual repellent point between desired locations. However, the DCM only presents a simplified version of the LIP dynamics, potentially limiting the control capabilities and versatility of footstep planning. Further to this, a reference trajectory for the DCM must be specified, which is generated heuristically \cite{Khadiv2016}.

Another approach is to optimize the full system dynamics, only considering the CoM state at each footstep. \cite{Kryczka2015} achieves this, interpolating between regenerated gait patterns when a disturbance is detected. \cite{Ding2019} also only considers the CoM state at each step, but reformulates the nonlinear optimization problem to a sequential quadratic programming problem by writing the nonlinear terms directly as decision variables. However, although \cite{Ding2019} is fast, the decision variables, including footstep positions require references to track - this again requires heuristics provided by the algorithm designer and may not represent the optimal behaviour.

Rather than simplifying the dynamics of the system, an alternative method is to simplify the optimization itself. Such approaches focus on generating approximate solutions that are iteratively improved with each control step. Real Time Iteration (RTI), introduced in \cite{Diehl2002} is an online method for approximately solving MPC problems, where the solution is found by first linearising the system at its current state, then performing one QP step \cite{Diehl2005}. RTI has been shown to significantly reduce computation time when solving nonlinear MPC problems \cite{Gros2020}. \cite{DBLP:journals/corr/abs-1712-02889} followed this approach and applied nonlinear MPC to control a quadruped robot's dynamic motion. However, their optimization is unconstrained and requires careful engineering. This has been improved recently by \cite{mastalli2020crocoddyl} with adding 'soft' constraints - penalty terms included in the cost function to shape it such that its' optima satisfy constraints. Constraints can more concretely be handled by using the augmented Lagrangian (AL) method \cite{20000922074}, which `augments' the standard Lagrange formulation with an added quadratic component. This method has gained popularity in robotics due to its fast convergence and numerical robustness \cite{howell2019altro}.

Terrain-aware (also known as terrain-adaptive) locomotion has recently started attracting growing attention. \cite{wu2010terrain} has achieved terrain-aware real-time locomotion for a 3D simulated humanoid with a gait library generated offline. The gait library is parameterized so that the desired pose planned in real-time based on the height map can choose the right gait to reach the goal. \cite{20.500.11850/425596} has demonstrated terrain-aware walking on the quadruped robot ANYmal by tightly integrating terrain perception for foothold planning. By generating the elevation map in real time and integrating the map's information into the optimization for the footholds, This method enabled ANYmal to climb stairs and obstacles of heights up to 33 \% of the robot’s leg length. \cite{pmlr-v100-xie20a, doi:10.1177/0278364919859425} have shown impressive results of uneven terrain walking with Cassie robot. However, the controller did not use perception to complete a locomotion task.
\begin{figure}[!htb] 
\centering 
\hspace*{-2cm}
\includegraphics[trim={1cm 0cm 1cm, 0cm},clip,scale=0.5]{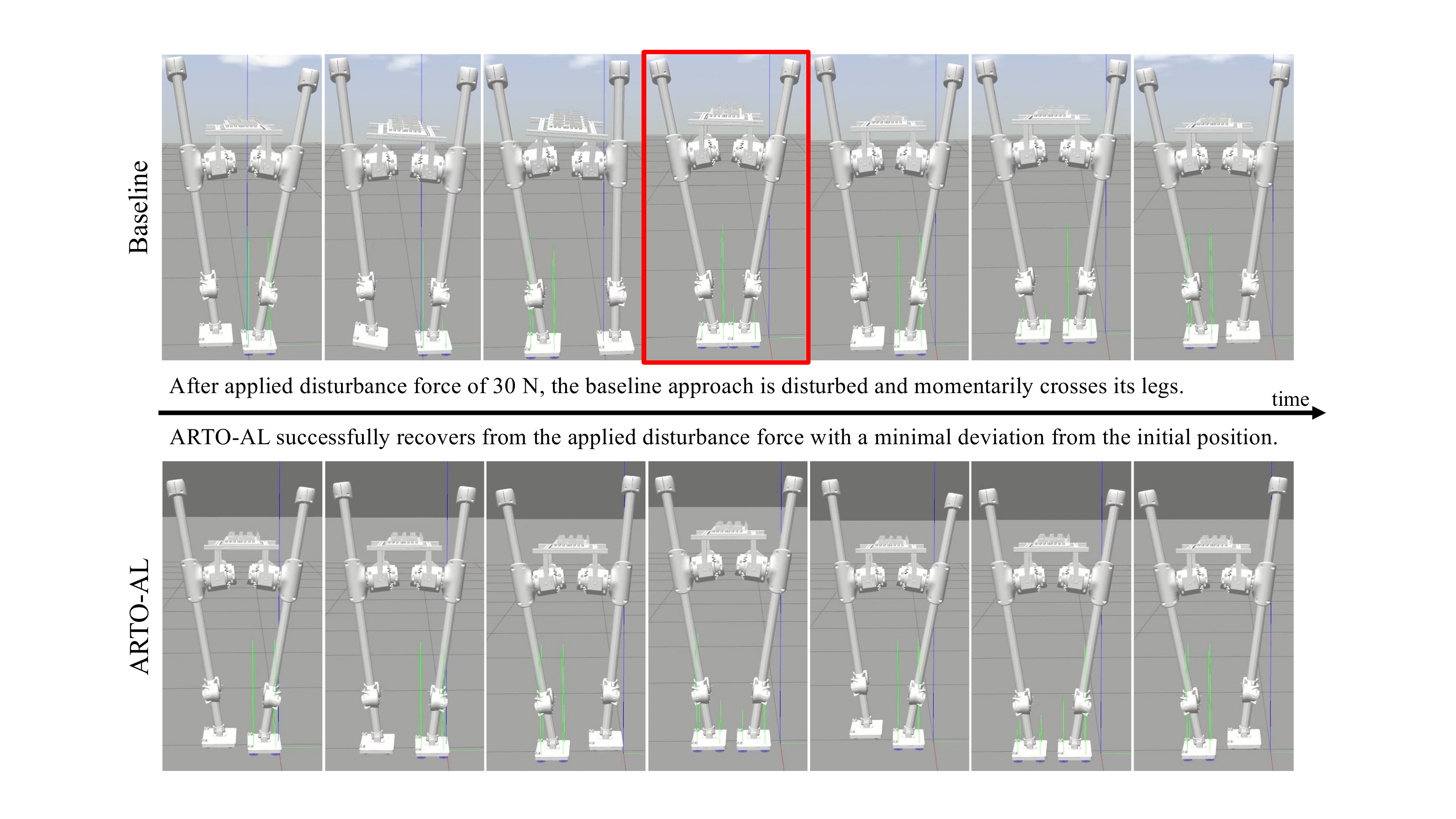}
\caption{With the baseline approach which does not optimize step timing, SLIDER suffers from the issue of crossing its legs after experiencing a big lateral push. While with ARTO-AL, SLIDER is able to successfully stabilise itself with a small deviation from the initial position after experiencing the same push.} 
\label{fig:frontImage} 
\end{figure}

In this work we investigate the use of the augmented Lagrangian method to solve the planning of footstep location and timing as a nonlinear MPC and demonstrate the robustness of the proposed approach to disturbances, as shown in Fig.\ref{fig:frontImage}. Our contributions are as follows:

\begin{itemize}
    \item We propose a footstep planner that adapts both footstep location and timing in real time. Instead of using simplified dynamics or doing local approximation, our optimizer uses analytical gradients of the full LIP model, allowing each gradient descent step to be computed in sub-millisecond time, resulting in a fast and robust footstep planner. 
    \item We propose a structure that combines the best of two optimisers asynchronously: the secondary interior point optimizer produces a near-optimal solution at a slower rate, which is used to initialise the main optimizer such that the solution can be quickly updated. This improves robustness of the planner.
    \item We prove how the dynamics used in the planner can be extended to three dimensions under mild assumptions, allowing real-time footstep planning to be performed on uneven terrains.
\end{itemize}

\section{PRELIMINARY}

\subsection{Linear Inverted Pendulum Model}

The dynamics of bipedal walking are commonly modelled using the Linear Inverted Pendulum (LIP) model. This model consists of an inverted pendulum that has its entire mass located at the top end of the pendulum with a rotation occurring at the lower end \cite{kajita3D}. The LIP model requires that the position of the lower end is constantly updated so that the Center of Mass (CoM) of the pendulum is stable. The additional constraint of the LIP model is that the CoM remains in the same plane at all times, maintaining a constant distance to the ground:

\begin{equation}\label{eq:lipx}
    \ddot{x}=\frac{g}{h}\left(x-u_{x}\right)
\end{equation}
\begin{equation}\label{eq:lipy}
    \ddot{y}=\frac{g}{h}\left(y-u_{y}\right).
\end{equation}
This can be equivalently written in state space form for both the sagittal and coronal motion, with $\bm{x} = [x,\, y,\, \Dot{x},\, \Dot{y}]^T$, and $\bm{u} = [u_x,\, u_y]^T$:
\begin{equation}
    \Dot{\bm{x}}
    =
    \begin{bmatrix}
        0   & 0     & 1     & 0 \\
        0   & 0     & 0     & 1 \\
        g/h & 0     & 0     & 0 \\
        0   & g/h   & 0     & 0
    \end{bmatrix}
    \bm{x}
    +
    \begin{bmatrix}
        0 & 0\\
        0 & 0\\
        -g/h & 0\\
        0 & -g/h
    \end{bmatrix}
    \bm{u}
    \label{eq:ssLIP}
\end{equation}
The full analytical solution to this second-order differential equation (\ref{eq:ssLIP}) can be expressed as an exponential equation (\ref{eq:exp}). Here we define $\Delta t_k$ as the step duration, $x_k$ and $\Dot{x}_k$ as the CoM position and velocity at the beginning of footstep $k$, and $u_{x,k}$ as the position of the foot.

\begin{equation}\label{eq:exp}
    x_{k+1} = a_{k}\exp(\omega \Delta t_k) + b_{k}\exp(-\omega \Delta t_k) + u_{x,k}.
\end{equation}
Where $a_k = 0.5(x_k - u_{x,k} + \frac{\Dot{x}_k}{\omega})$, and $b_k = 0.5(x_k - u_{x,k} - \frac{\Dot{x}_k}{\omega})$ and $\omega=\sqrt{\frac{g}{h}}$. Note that the positive exponential term is linked to the divergent component of motion, as it diverges as $\Delta t_k$ increases. 

As can be seen, equation (\ref{eq:exp}) is non-linear with footstep duration, meaning real-time optimization is difficult. This is further exacerbated as the optimization horizon increases, as subsequent optimisations are recursively linked to their predecessors. For more detailed formulation of these relationship, please see the Appendix.

\subsection{Solving Constrained Optimisation with Augmented Lagrangian} \label{section:constraint}

Constrained optimization problems are commonly solved by converting the problems into unconstrained ones by using Lagrange multipliers (LM). These problems can be expressed in a general form as:
\begin{equation}
  \begin{aligned}
    \min_{x} f(x)  \\
    subject \; to \quad c(x)\leq 0
    \end{aligned}
    \label{eq3}
\end{equation}
with
\begin{equation}
     \;\; c(x)=\{\begin{array}{cl}
c_{j}(x) & j \in \mathrm{E} \\
\max \left[0, c_{j}(x)\right] & j \in \mathrm{I}.
\end{array}
\label{eq4}
\end{equation}

where $c(x)$ is the is the equality (E) or inequality (I) constraint, as shown in equation (\ref{eq4}). A popular way to solve the constrained optimization problem is perform dual gradient descent on the Lagrangian of the problem:
\begin{equation}
    \mathcal{L}=f(x)+\sum_{j=0}^{P} \lambda_{j} c(\mathrm{x})_{j}.
\end{equation}

where $\lambda_{j}$ is the $j$th Lagrange multiplier. As $\lambda$ becomes larger, the violation of the constraints is penalised more severely, forcing the optimizer to find a solution within the feasible set. This behaviour leads to a $\lambda$ that updates at each iteration to achieve a value of infinity in order to obtain the optimal solution, however an infinite coefficient would cause an infinite penalty when the constraint is violated \cite{bertsekas2014constrained}. As shown in \cite{NoceWrig06}, an improved method of optimization is to use the Augmented Lagrangian method, which adds a quadratic penalty term to the Lagrangian:
\begin{equation} \label{eq:AL}
    \mathcal{L}=f(x)+\sum_{j=0}^{P} \lambda_{j} c(\mathrm{x})_{j}+\sum_{j=0}^{P} \mu c^{2}(\mathrm{x})_{j}.
\end{equation}

where $\mu$ is the positive coefficient of the AL penalty term. The solutions and coefficients of Augmented Lagrangian method are iteratively updated using gradient descent as follows:
\begin{align} \label{eq:ARTO:update_1}
    d x_{k+1}=d x_{k}-\alpha \frac{d L}{d x}\\
\label{eq:ARTO:update_2}
    \lambda_{k+1}=\lambda_{k}+\mu_{k} c(x)\\
\label{eq:ARTO:update_3}
   \mu_{k+1}=\varphi \mu_{k} \quad with \;\; \varphi>1.
\end{align}

where $x$ is the solution to the optimization problem, $\alpha$ is the update step size, and $\psi$ is a scaling parameter that regulates the update of $\mu$ by monotonically increasing it.

\section{METHODS}
\subsection{System Overview}
\begin{figure}[!htb]
\centering 
\includegraphics[trim={6cm 3cm 5cm, 3cm},clip, scale=0.6]{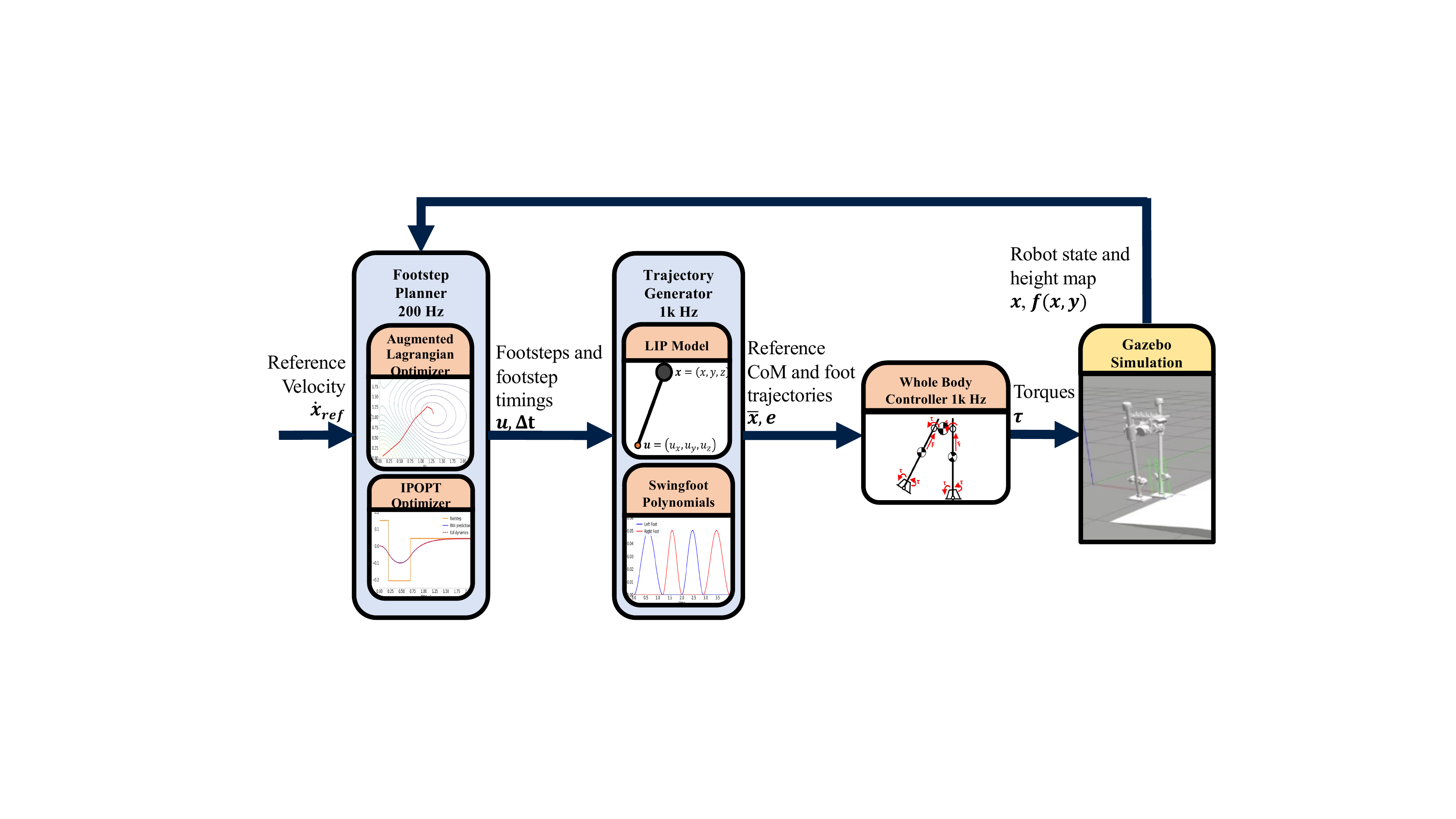}
\caption{Illustration of the terrain-aware footstep planner pipeline. Given the reference velocity from the operator, robot states and the height map representing terrain information, the planner outputs reference footstep location in real time.} 
\label{fig:structure_overview} 
\end{figure}
The proposed footstep planner solves the optimal control problem introduced in equation (\ref{eq:footstep_planning}), producing footstep locations and durations at a rate of $200$~Hz. As summarized in Fig. \ref{fig:structure_overview}, given a reference CoM velocity, the footstep planner makes use of two optimizers to generate optimal solutions. The main optimizer makes use of the AL method to rapidly update the footstep plan at $200$~Hz, while a secondary interior point optimizer produces more optimal solutions at 20~Hz. From the generated footstep plan, a CoM trajectory is generated following the LIP dynamics and a swing foot trajectory is generated using fifth order polynomials. Finally, an inverse dynamics-based whole body controller tracks the generated trajectories and outputs joint torques at a frequency of 1~kHz.

\subsection{Footstep Location and Timing Optimization Formulation}
Footstep planning for bipedal walking can be formulated as an optimal control problem. A cost function is specified to allow the CoM state $\bm{x}_k$ to follow a reference $\bm{x}_{\text{ref}}$. At each control step $k$, this cost is minimised by two control inputs: footstep location $\bm{u}_k$, and the duration of the step $\Delta t_k$. This optimisation is constrained by the CoM dynamics which are not only dependent on the robot's physical characteristics, but also on the model used to represent them. Furthermore, there is a set of physical inequality and equality constraints that this cost function is subject to:
\begin{align}\label{eq:footstep_planning}
            \min_{\bm{u}_{k:N},\,\Delta t_{k:N}}\sum_{k=0}^{N}||\bm{x}_k - \bm{x}_{\text{ref}}||_Q^2\\
            \text{s.t.}\\
            \bm{x}_{k+1} = \bm{f}(\bm{x}_k, \Delta t_k, \bm{u}_k)\\
            \bm{g}(\bm{x}_k, \Delta t_k, \bm{u}_k) \leq \bm{0}\\
            \bm{h}(\bm{x}_k, \Delta t_k, \bm{u}_k) = \bm{0}.
\end{align}

\begin{figure}[!htb] 
\centering 
\includegraphics[scale=1.0]{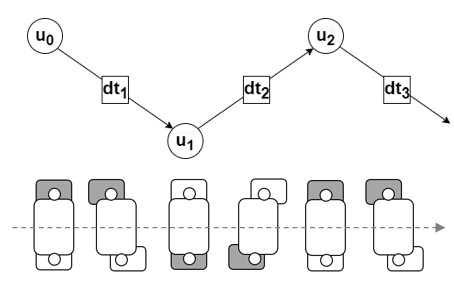}
\caption{Illustration of the prediction horizon. The current footstep position of the active supporting foot (grey area) is represented by $u_0$, which will switch to the next footstep locations $u_1$  and $u_2$ in the duration $dt_1$, $dt_2$, and $dt_3$.} 
\label{fig:step_illustration} 
\end{figure}

The prediction horizon in this formulation is finite. Importantly, in this work we consider each footstep as a single control step, as opposed to considering the centre of pressure of the foot during each step. This greatly simplifies the optimisation problem while allowing us to plan further into the future. In this work, we consider a prediction horizon of two footsteps ahead: optimizing two footstep positions and three footstep durations (including the current footstep duration). The footstep positions and durations optimized can be illustrated graphically in Fig.\ref{fig:step_illustration}.
At each control step, each optimizer must produce optimal footstep locations and durations. A reference CoM trajectory is defined by a reference velocity, such that weighting matrix $Q = [w_x\ 0;\ 0\ w_y]$.
The dynamic equality constraint defined by the linear inverted pendulum model defined in the previous equations \ref{eq:exp} and physical inequality constraints are added:
\begin{align}\label{eq:formulation}
\left\|\boldsymbol{x}_{p o s, k}-\boldsymbol{u}_{k}\right\|_{2} \leq l_{\max }, \tag*{(next step length)}\\
    \left\|\boldsymbol{x}_{p o s, k}-\boldsymbol{u}_{k-1}\right\|_{2} \leq l_{\max }, \tag*{(current step length)}\\
    \left\|\boldsymbol{x}_{vel, k+1}\right\|_{2} \leq v_{\max }, \tag*{(velocity limit)}\\
     \left|\boldsymbol{u}_{k, y}-\boldsymbol{u}_{k-1, y}\right| \geq r_{\text {foot }}, \tag*{(no crossing feet)}\\
     c_{t, k, \text { lower }} \leq \Delta t_{k} \leq c_{t, k, \text { upper }}. \tag*{(step time)}
\end{align}

Step length constraints define a maximum leg extension such that the robot does not travel too far on its support foot, and doesn't place the next foot too far away. No crossing feet constraint prevents the current and next support foot from crossing each other in the coronal plane. Time limit constraint defines safety limits to the step duration.

\subsection{Augmented Lagrangian Optimizer}
The AL optimizer makes use of analytically calculated gradients in order to find approximate optima as quickly as possible. However, gradient methods often have difficulty handling constraints, so the AL method introduced in Section \ref{section:constraint} is used. The analytical gradient descent is computed by taking the derivative of equation (\ref{eq:footstep_planning}) with respect to the x-y footstep location $\bm{u}_k$, shown in equation (\ref{eq:ARTO:AL_derivative_1}), and footstep duration $\Delta t_k$, shown in equation (\ref{eq:ARTO:AL_derivative_2}):

\begin{equation} \label{eq:ARTO:AL_derivative_1}
    \frac{\partial \mathcal{L}}{\partial \bm{u}_k} = 2 \sum^{N}_{k=1} \frac{\partial \bm{x}_k}{\partial \bm{u}_k}\mathbf{Q}(\bm{x}_k-\bm{x}_\text{ref}) + \frac{\partial}{\partial \bm{u}_k} \sum^{P}_{j=1} \lambda_{j}c(\bm{x})_{j} + \frac{\partial}{\partial \bm{u}_k} \sum^{P}_{j=1}c(\bm{x})^{T}_{j}\mu_j c(\bm{x})_j
\end{equation}

\begin{equation} \label{eq:ARTO:AL_derivative_2}
    \frac{\partial \mathcal{L}}{\partial \Delta t_k} = 2 \sum^{N}_{k=1} \frac{\partial \bm{x}_k}{\partial \Delta t_k}\mathbf{Q}(\bm{x}_k-\bm{x}_\text{ref}) + \frac{\partial}{\partial \Delta t_k} \sum^{P}_{j=1} \lambda_{j}c(\bm{x})_{j} + \frac{\partial}{\partial u_{k}} \sum^{P}_{j=1}c(\bm{x})^{T}_{j}\mu_j c(\bm{x})_j.
\end{equation}
Individual expressions for the derivatives of CoM position $x_k$ and velocity $\dot{x}_k$ can be found in the Appendix.

We define the criteria of convergence is that the difference for norm of gradients is smaller than 0.05. In the case that the solution is outside the bounds defined by constraints, the footstep planner algorithm uses the projection method to “project” the solution on the lower and upper bounds of constraints, this can also prevent oscillations observed during the convergence. One special case is the remaining time of current step, instead of projecting to boundaries, remaining time is subtracted by the duration of one iteration, as shown in equation (\ref{eq:projection}):
\begin{equation}\label{eq:projection}
    dt_{0,k+1} = dt_{0,k} - 1/\text{LoopRate},\quad \text{if} \; dt_0 < 0.
\end{equation}
where $k$ represents the current iteration and $k+1$ represents the next iteration.

\subsection{Interior Point Optimizer}
The secondary optimizer uses Interior Point methods to perform the optimization. In this work we use the IPOPT optimizer \cite{Wachter2006} via the software framework CasADi \cite{Andresson2012}. This optimizer is considerably slower, but more accurate than the AL optimizer. To warm start the solver, the initial value of the solution, $\bm{u}_{k,\,j}^{(i)}$, is set as the previous iteration's final solution, $\bm{u}_{k,\,j-1}^{(f)}$, for each step in the prediction horizon, $k \in [1, ..., N]$. The current step duration, $\Delta t_{1,\,j}^{(i)}$ is adjusted for the time elapsed between iterations, $\tau_{j-1}$:
\begin{align}
    \bm{u}_{k,\,j}^{(i)} = \bm{u}_{k,\,j-1}^{(f)}\\
    \Delta t_{1,\,j}^{(i)} = \Delta t_{1,\, j-1}^{(f)} - \tau_{j-1}.
\end{align}
When a step has been taken between iterations, the initial value of the solution is provided as follows for the first $k \in [1, ..., N-1]$ predicted steps:
\begin{align}
    \bm{u}_{k,\, j}^{(i)} = \bm{u}_{k+1,\, j-1}^{(f)}\\
    \bm{\Delta t}_{k,\,j}^{(i)} = \bm{\Delta t}_{k+1,\, j-1}^{(f)}.
\end{align}
With the final predicted step initialized to repeat the motion of the penultimate step, mirrored in the $y$ direction.
\begin{align}
    \bm{u}_{N,\, j}^{(i)} = \bm{u}_{N-1,\, j-1}^{(i)} + 
    \begin{bmatrix}
        1   & 0     \\
        0   & -1    \\
    \end{bmatrix}
    ( \bm{u}_{N,\, j-1}^{(i)} - \bm{u}_{N-1,\, j-1}^{(i)} )\\
    \Delta t_{N,\,j}^{(i)} =  \Delta t_{N,\,j-1}^{(i)}.
\end{align}

\subsection{Asynchronous Optimization}
\begin{figure}[!htb]
\centering 
\includegraphics[trim={1cm 0cm 1cm, 0cm},clip,scale=0.5]{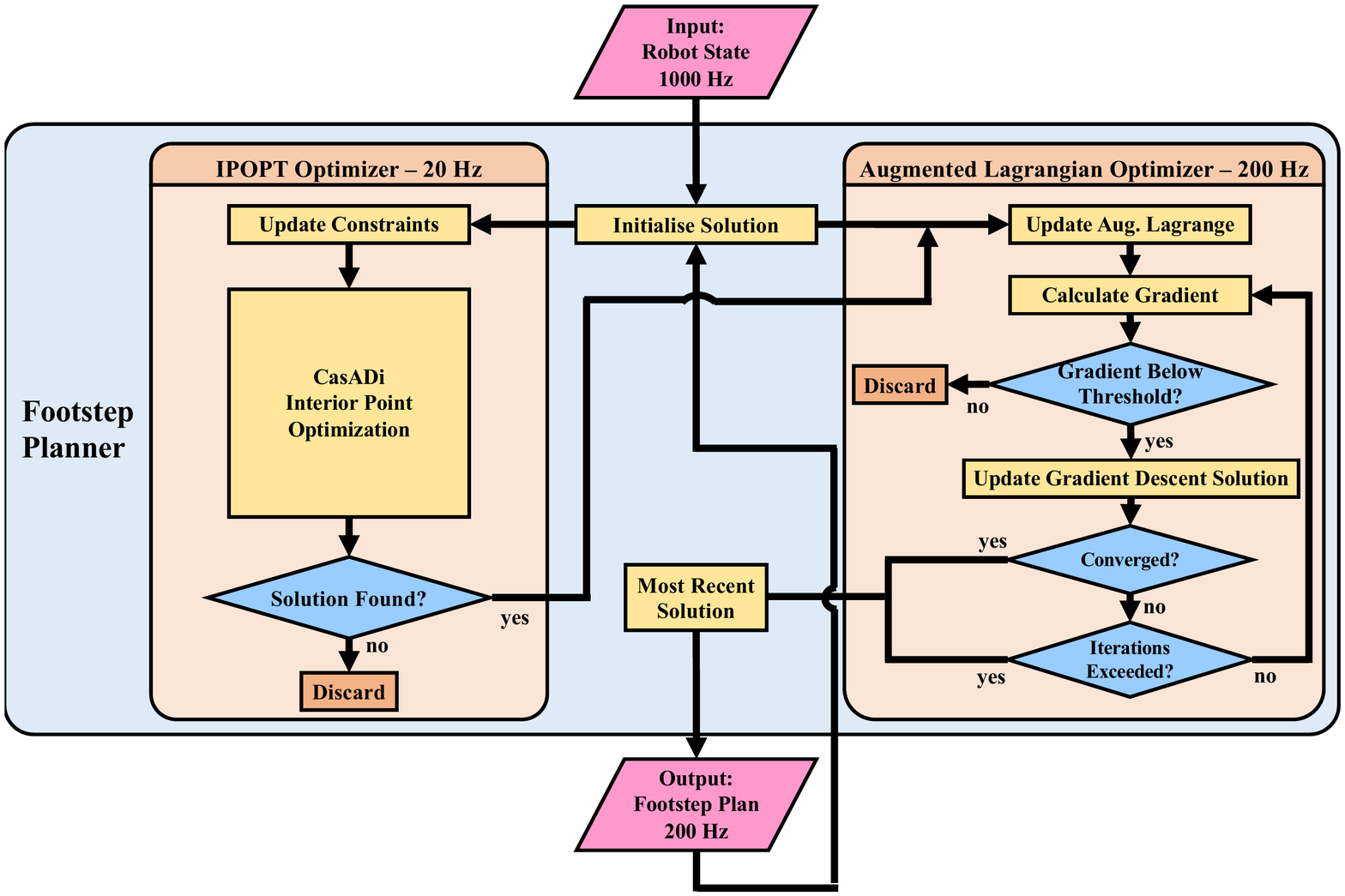}
\caption{Flowchart describing the asynchronous optimization process.} 
\label{fig:ARTO:ARTO_structure} 
\end{figure}

The AL and Interior Point optimizers run in parallel threads communicating via ROS topics. Each time a feasible solution is computed from IPOPT, the solutions, as well as the values of Lagrange multipliers, are fed into the gradient descent optimizer to reinitialize it, as shown in Fig.\ref{fig:ARTO:ARTO_structure}. Importantly, we observe that the AL optimizer alone can get stuck in local solutions, or in some situations the analytical gradient is extremely large (for example at the switching point between feet), meaning the AL optimizer's solution is not feasible. For periods of time where an AL optimizer solution is infeasible, the IPOPT solution is used as a warm-start for the AL optimizer to generate feasible solutions. The interaction between the solvers is shown in Fig.\ref{fig:ARTO:ARTO_structure}. This method allows the footstep planner to run at a rate of $200\,$Hz: an order of magnitude increase of speed over the interior point optimizer alone. This structure combines the best of both optimizers: IPOPT has the advantage of finding accurate optima, and the AL optimizer is very fast.

\subsection{Terrain-Aware Footstep Planning}

\begin{figure}[!htb] 
\centering 
\includegraphics[width=0.8\textwidth]{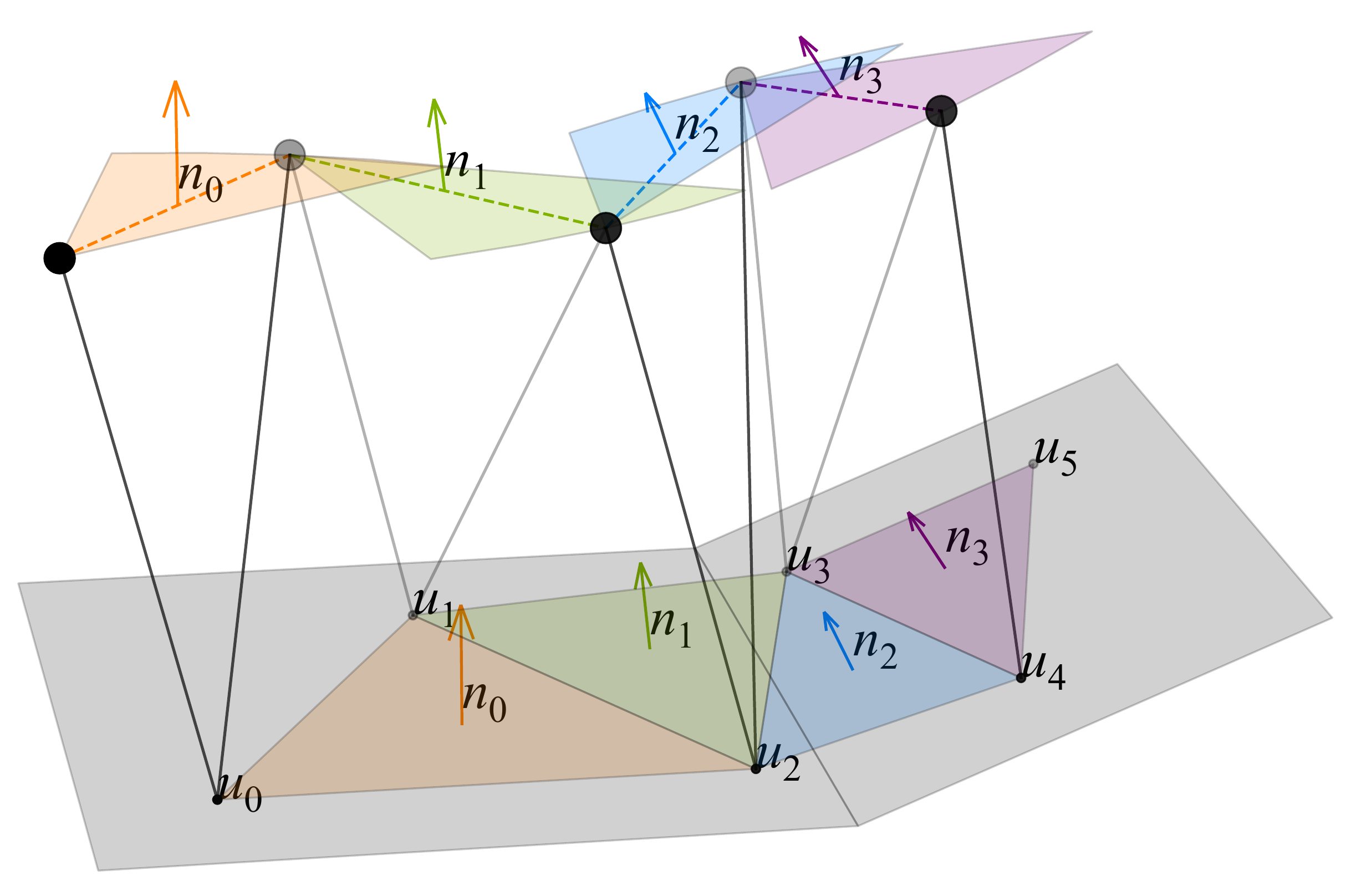}
\caption{3D illustration of constructing the plane that the CoM moves along from optimal footsteps.} 
\label{fig:ARTO:slope} 
\end{figure}

Inspired by the parallel height constraint of the LIP model proposed by \cite{kajita20013d}, the LIP model can be actually extended as a viable model for dealing with uneven terrain by constraining the CoM to move along a plane parallel to the slope of the terrain, as shown in Fig.\ref{fig:ARTO:slope}. Because the plane has a linear relationship with the $x$ and $y$ position of the robot, no new decision variables are added to the optimization problem.

If we consider the dynamics of an inverted pendulum, its motion in the sagittal ($x-z$) plane is described by the following equation:
\begin{equation} \label{eq:IP_sagittal}
    \frac{\Ddot{x}}{x-u_x} = \frac{\Ddot{z}+g}{z-u_z},
\end{equation}

and in the coronal ($y-z$) plane:
\begin{equation} \label{eq:IP_coronal}
    \frac{\Ddot{y}}{y-u_y} = \frac{\Ddot{z}+g}{z-u_z}.
\end{equation}

We assume that the local terrain around the robot can be described as a plane defined by gradients in the $x$ and $y$ directions of $\alpha$ and $\beta$, respectively. The motion of the centre of mass is then constrained to move parallel to this plane at a vertical offset (height) of $h_0$:
\begin{equation}\label{eq:IP_plane}
    z-u_z = \alpha(x-u_x) + \beta(y-u_y) + h_0,
\end{equation}
The acceleration in the $z$ direction can then be defined as:
\begin{equation}\label{eq:IP_plane_acc}
    \Ddot{z} = \alpha\ddot{x} + \beta\ddot{y},
\end{equation}

This constraint leads to (\ref{eq:IP_sagittal}) and (\ref{eq:IP_coronal}) becoming non-linear, and difficult to optimise. However, in two cases, they can be reduced to the popular linear inverted pendulum equations.

We start by substituting (\ref{eq:IP_plane}) into (\ref{eq:IP_sagittal}) and (\ref{eq:IP_coronal}), giving

\begin{align}
    \frac{\Ddot{x}}{x-u_x} = \frac{\alpha\Ddot{x}+\beta\Ddot{y}+g}{\alpha(x-u_x) + \beta(y-u_y) + h_0},\label{eq:IP_xplane}
    \\
    \frac{\Ddot{y}}{y-u_y} = \frac{\alpha\Ddot{x}+\beta\Ddot{y}+g}{\alpha(x-u_x) + \beta(y-u_y) + h_0}.\label{eq:IP_yplane}
\end{align}

The first case in which these equations reduce to the linear inverted pendulum model, as noted in \cite{Kajita2001}, is when $\alpha$ and $\beta$ are very small (i.e. flat ground). This is of little interest to a robot walking on uneven terrain.

The second case occurs when the forward ($x$) direction of the robot aligns with the direction of steepest gradient of the plane. This reduces the gradient in the $y$ direction to zero, namely $\beta=0$. Then (\ref{eq:IP_xplane}) reduces to the LIP equation. Next, we assume that the distance between the centre of mass and the foot in the $y$ direction is significantly larger than in the $x$ direction, due to the inter-foot spacing of a bipedal robot:
\begin{equation}\label{eq:IP_assumption_y>x}
    y - u_y \gg x - u_x.
\end{equation}
Finally, we assume that the acceleration of the sway of the robot side-to-side is much greater in magnitude than the robot's forward acceleration:
\begin{equation}\label{eq:IP_assumption_y>x_dot}
    \Ddot{y} \gg \Ddot{x}.
\end{equation}

Under these assumptions, the $x$ terms from (\ref{eq:IP_yplane}) are negligible, meaning that it, too, reduces to the equation of a linear inverted pendulum.

The linear inverted pendulum dynamic model can therefore be applied to a walking robot on uneven terrain, provided that the robot is able to perceive the terrain and the robot is aligned with the direction of the local gradient of the terrain. This is the case for many structured environments, such as stairs and ramps.

The mapping of the terrain is achieved by adapting the state-of-the-art Robot-Centric Elevation Mapping ROS package provided by ANYbotics \cite{Fankhauser2018ProbabilisticTerrainMapping, Fankhauser2014RobotCentricElevationMapping} to our robot. This provides multiple 2.5d probabilistic mappings of the environment with information on the height, normals, variance and corresponding uncertainties. 
This mapping is updated based on new observations with the uncertainty being determined by the time of the most recent observation. The choice of a robot centric representation allows for planning on a local level when only local knowledge is necessary while not compromising on the possibility of global planning with reliable odometry estimates.

With these assumptions, we can construct a controller that computes the control inputs based on the LIP dynamics, while it is able to adapt to uneven terrain using perceptive feedback. The movement planes are constructed using a height map, which provides heights for queried sets of points. The queried points are centered around the locations of the next 2 footsteps and contain the averaged value of all points within the size of a footstep, as computed by the model predictive controller shown in Fig.\ref{fig:ARTO:slope}. These queried points and their heights are then used to fit a plane passing through them, and are passed to the controller which moves the CoM onto those planes. The planes can get updated mid-step without affecting performance or walking stability. The orientation of the swing foot is also commanded to align with the normals queried around the footstep position. This allows the robot to walk around increasingly complex terrains where the orientation of the CoM plane is different than the orientation of the individual footholds. In this paper, we update the plane at a rate of 5~Hz.

\subsection{Swing Foot Trajectory Generation}

The swing foot trajectory is generated using fifth order polynomials: current and final positions, velocities and accelerations are specified, and a parametric quintic curve in the $x$, $y$, and $z$ directions is generated to produce smooth trajectories. The $z$ polynomial is generated in two halves, with a midpoint foot height with zero velocity and acceleration specified to ensure no collision with the ground. The LIP model is used to generate the CoM trajectory during footsteps. Each trajectory point is calculated at the whole body controller rate of $1000\,$ Hz.

\subsection{Whole Body Controller}
We use an inverse dynamics-based whole body controller to track the foot motion while respecting a set of constraints. In the paper, the tasks of interest are the CoM position and velocity, the pelvis orientation, the foot positions and orientations. Each task is comprised of a desired acceleration as a feed-forward term and a state feedback term to stabilize the trajectory. Generally, the task for the linear motion can be expressed as:
\begin{equation*}
    \bm{J}_{\mathrm{T}}\ddot{\bm{q}} =  \ddot{\bm{x}}^{\mathrm{cmd}} - \dot{\bm{J}}_{\mathrm{T}}\dot{\bm{q}},
\end{equation*}
\begin{equation*}
    \ddot{\bm{x}}^{\mathrm{cmd}} = \ddot{\bm{x}}^{\mathrm{des}} + \bm{K}_{\mathrm{P}}^{\mathrm{pos}}(\bm{x}^{\mathrm{des}} - \bm{x}) + \bm{K}_{\mathrm{D}}^{\mathrm{pos}}(\dot{\bm{x}}^{\mathrm{des}} - \dot{\bm{x}}),
\end{equation*}
where $\bm{J}_{\mathrm{T}}$ is the translational Jacobian for the task, $\bm{x}$ is the actual position of the link, and the superscript $\mathrm{des}$ indicates the desired motion. 

For the task of angular motion, the command can be formulated as:
\begin{equation*}
    \bm{J}_{\mathrm{R}}\ddot{\bm{q}} =  \dot{\bm{\omega}}^{\mathrm{cmd}} - \dot{\bm{J}}_{\mathrm{R}}\dot{\bm{q}},
\end{equation*}
\begin{equation*}
    \dot{\bm{\omega}}^{\mathrm{cmd}} = \dot{{\bm \omega}}^{\mathrm{des}} + \bm{K}_{\mathrm{P}}^{\mathrm{ang}}(\mathrm{AngleAxis}(\bm{R}^{\mathrm{des}}\bm{R}^{\mathrm{T}})) +  \bm{K}_{\mathrm{D}}^{\mathrm{ang}}(\bm{\omega}^{\mathrm{des}} - \bm{\omega}),
\end{equation*}
where $\bm{J}_{\mathrm{R}}$ is the rotational Jacobian for the task, $\bm{R}$ and $\bm{R}^{\mathrm{des}}$ denote the actual and desired orientation of the pelvis link respectively, 
$\mathrm{AngleAxis}()$ maps a rotation matrix to the corresponding axis-angle representation to avoid the gimbal lock of using euler angles, 
$\bm{\omega} \in \bm{R}^3$ is the angular velocity of the link.

\subsubsection{QP formulation}
Inspired by \cite{Herzog_2015}, the full dynamics of the walking robot can be decomposed into the underactuated part and actuated part:
\begin{equation*}
    \begin{bmatrix}
    \bm{M}_f \\ \bm{M}_a 
    \end{bmatrix} \bm{\ddot{q}} +
    \begin{bmatrix}
        \bm{H}_f \\ \bm{H}_a
    \end{bmatrix} = 
    \begin{bmatrix}
        \bm{0} \\ \bm{S}_a
    \end{bmatrix} \bm\tau +
    \begin{bmatrix}
        \bm{J}^{\mathrm{T}}_f \\ \bm{J}^{\mathrm{T}}_a
    \end{bmatrix} \bm{f},
\end{equation*}
where $\bm{M}$, $\bm{H}$, $\bm{S}_a$, $\bm {\tau}$, $\bm{J}$ and $\bm{f}$ are the mass matrix, Coriolis force matrix and gravitation force vector, the actuator selection matrix, joint torques vector, the stacked contact Jocabian and reaction force vector. The subscript, $f$ and $a$, indicates the floating part and actuated part respectively. The weighted sum formulation is applied, in which one QP problem is solved at each control loop.
The formulation of the QP problem can be written as 
\begin{align}
\min_{\ddot{\bm{q}},\, \bm{f}} \quad & \frac{1}{2}\| \bm{A}\ddot{\bm{q}} + \dot{\bm{A}}\dot{\bm{q}} - \bm{B}^{\mathrm{cmd}} \|_{\bm{W}_1}^2\\
\textrm{s.t.} \quad & \bm{M}_f\bm{\ddot{q}} - \bm{J}^{\mathrm{T}}_f\bm{f} = - \bm{H}_f \tag*{(floating base dynamics) }\\
& \bm{P}\bm{f} \leq \bm{0}\tag*{(friction cone)}\\
& \bm{S}_a^{-1}(\bm{M}_a \ddot{\bm{q}} + \bm{H}_a - \bm{J}^{\mathrm{T}}_a\bm{f}) \in [\bm{\tau}_{min},\, \bm\tau_{max}]\tag*{(input limits)},
\end{align}

where $\bm{A}$ is a stack of the Jacobian matrices for the tasks of interest, $\bm{B}^{\mathrm{cmd}}$ is a stack of the commanded accelerations and $\bm{W}_i \, (i = 1, \,2)$ are the weighting matrices, $\bm{P}$ denotes the linearized friction cone matrix. We treated the unilateral contact constraint as a soft constraint by simply assigning a large weight on the desired zero acceleration. It is reported in \cite{feng2014optimization} that this gives a better stability.  

The output torque commands $\bm \tau$ at each control iteration is computed by
\begin{equation}
    \bm\tau = \bm{S}_a^{-1}(\bm{M}_a \bm{\ddot{q}} + \bm{H}_a - \bm{J}^{\mathrm{T}}_a\mathbf{f}).
\end{equation}
To aid stable walking on uneven terrains, a different set of gains is applied for the swing and the stance foot. The stance foot is generally stable and low gains are preferred. Adversely on the swing foot the trajectory is aimed to be accurately tracked and thus high gains are applied. 

\section{EXPERIMENTAL RESULTS}
\subsection{Experimental Platform and Scenarios}
\begin{figure}[!htb]
\centering
\includegraphics[width=0.55\columnwidth]{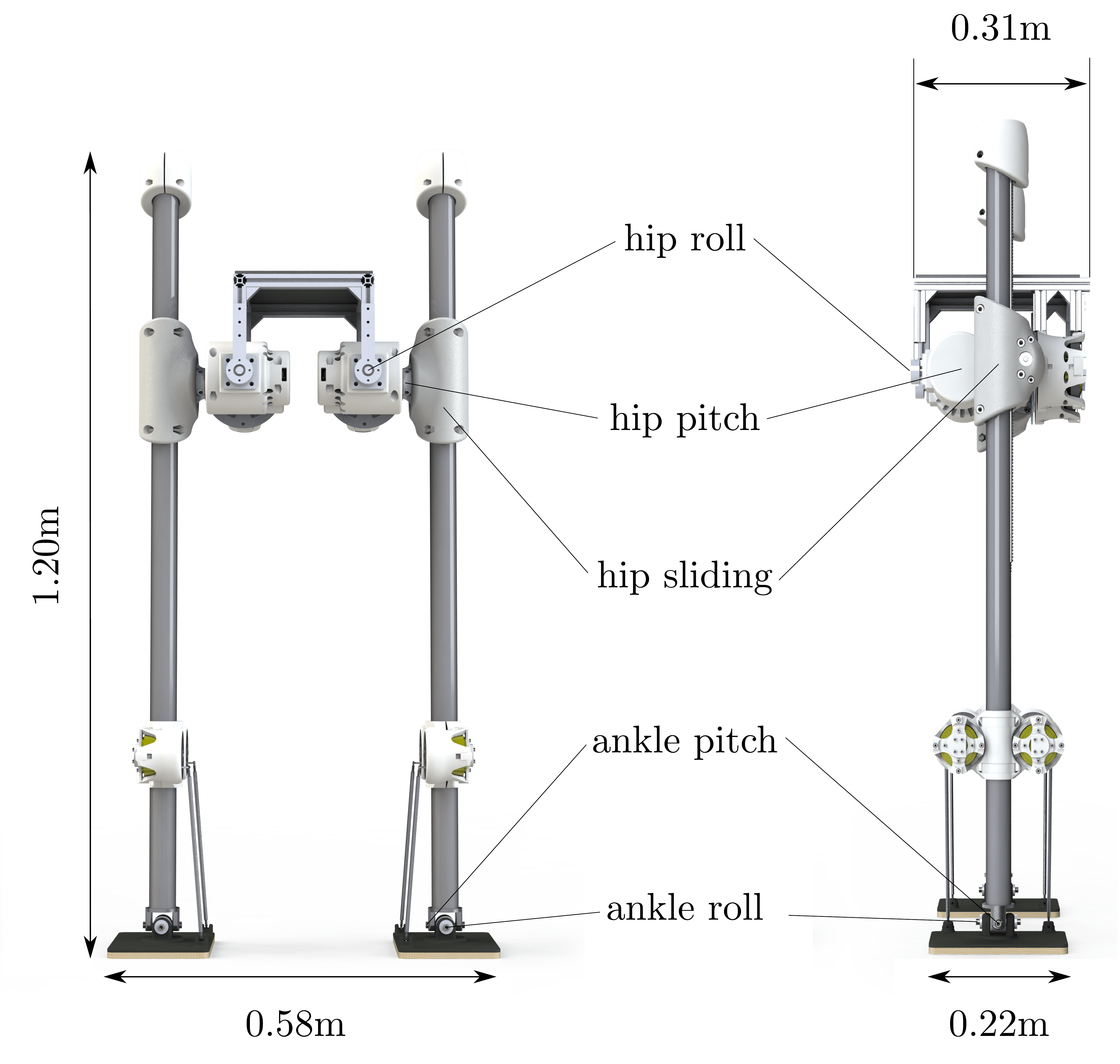}
\caption{The dimensions and joint configuration of SLIDER.}
\label{fig:slider_dim_config}
\end{figure}

We use the SLIDER robot as our experimental platform. {SLIDER} is a knee-less bipedal robot designed by the Robot Intelligence Lab at Imperial College London \cite{Wang2020}. As shown in Fig.\ref{fig:slider_dim_config}, SLIDER is 1.2~m tall and has 10 Degrees of Freedom (DoF); hip pitch, hip roll, hip slide, ankle roll and ankle pitch on each leg. The robot is lightweight (14~kg in total) and most of its weight is concentrated in the pelvis as its legs are made of carbon fiber reinforced polymer.
 The prismatic knee joint design is {SLIDER}'s unique feature that differentiates it from other robots with anthropomorphic design. The sliding joint has a large range of motion, which combined with the light weight makes {SLIDER} suitable for agile locomotion.  
 
\subsection{Flat Ground Walking}
\begin{figure}[!htb]
    \centering
    \includegraphics[scale=0.6, trim={7.5cm 5cm 7.5cm 5cm}, clip]{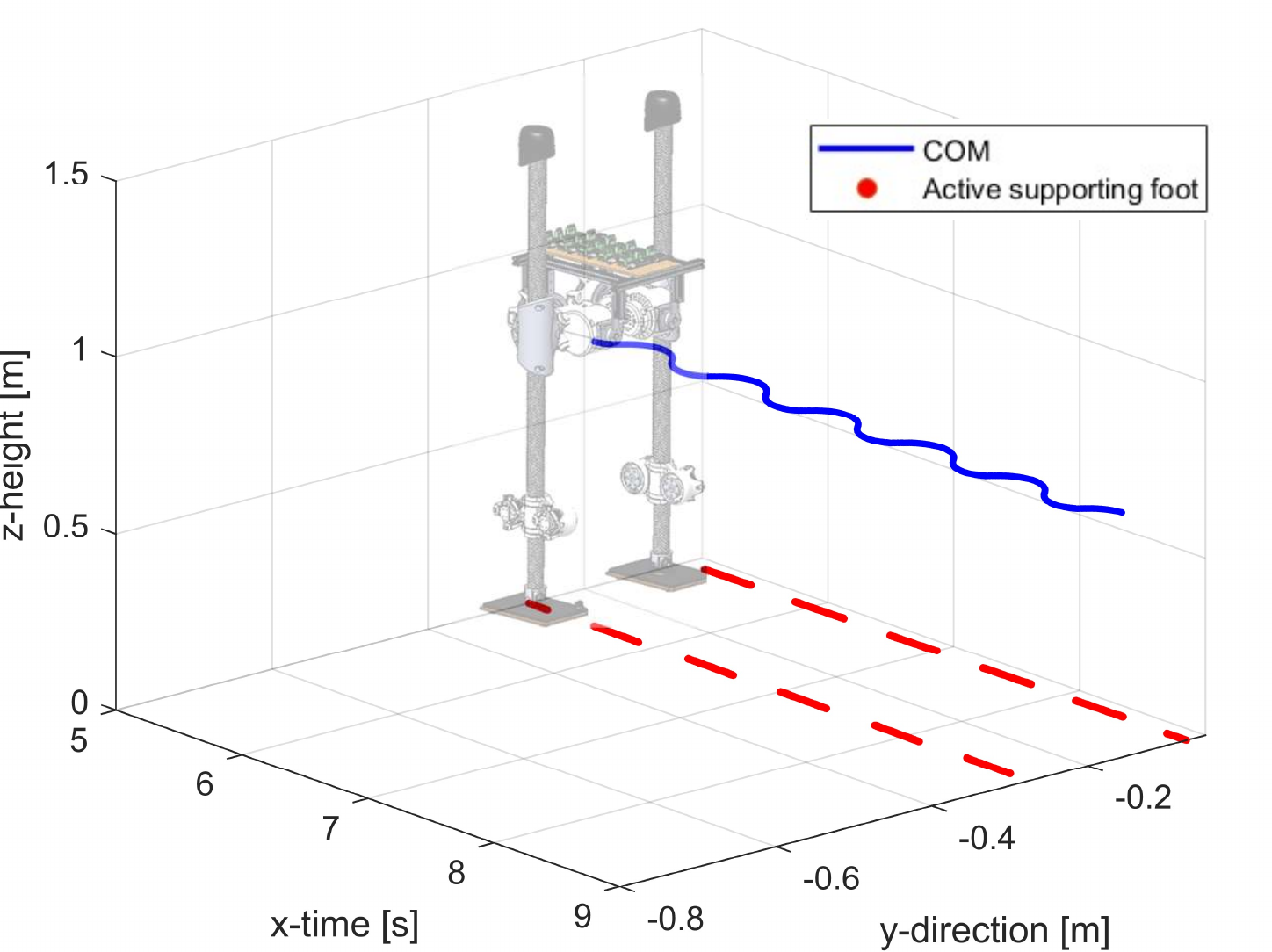}
    \caption{3D illustration of CoM (blue) and supporting foot position (red) when SLIDER is walking forward at 0.2 m/s with the planner ARTO-AL.}
    \label{fig:ARTO:flat_walking_1}
\end{figure}

\begin{figure}[!htb]
    \centering
    \includegraphics[scale=0.6, trim={7.5cm 5cm 7.5cm 5cm}, clip]{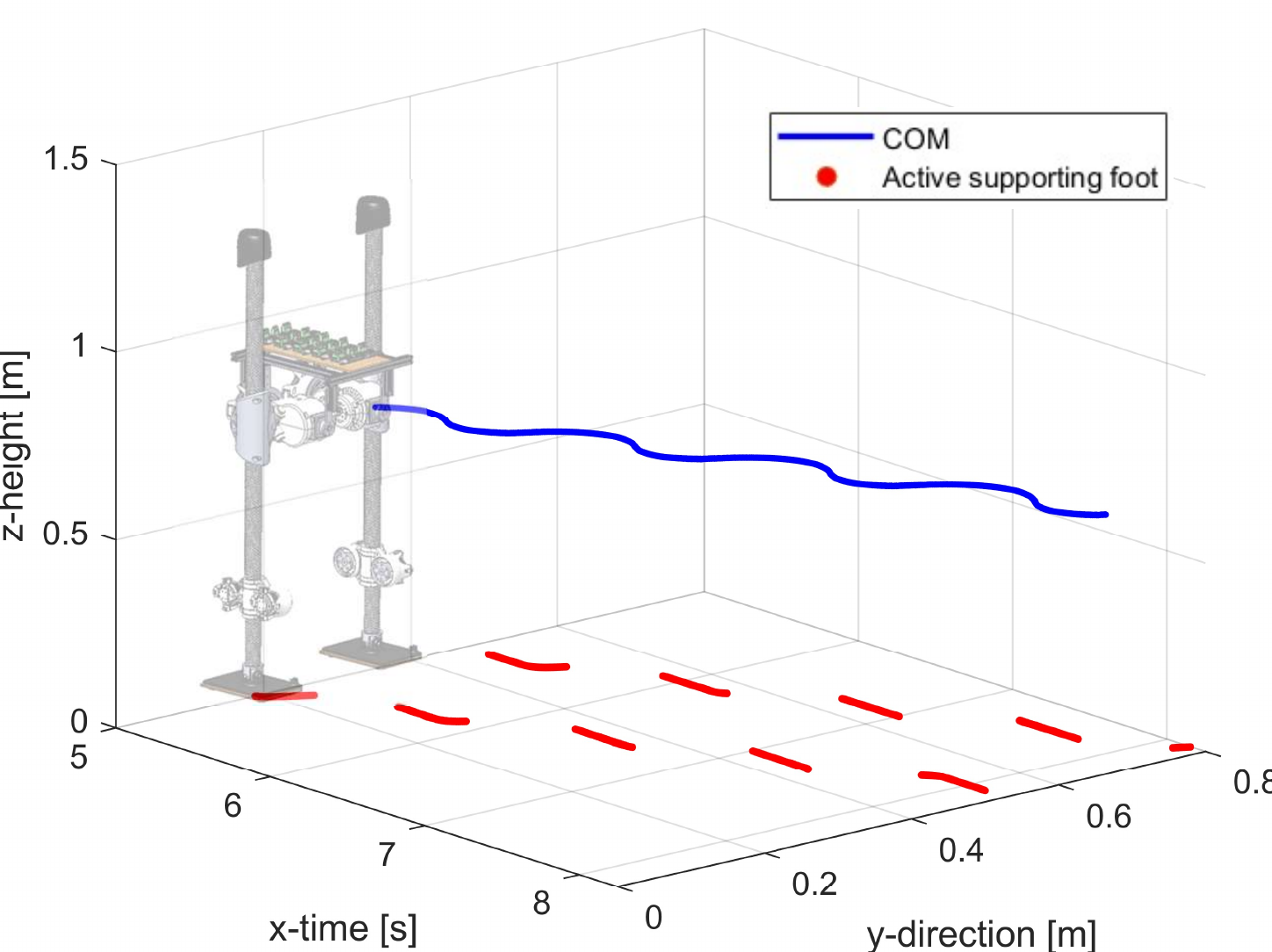}
    \caption{3D illustration of CoM (blue) and supporting foot position (red) when SLIDER is walking sideways at 0.3 m/s with ARTO-AL.}
    \label{fig:ARTO:flat_walking_2}
\end{figure}

SLIDER successfully walked 50 m of simulation ground at forward speed of 0.2 m/s in Gazebo. Fig.\ref{fig:ARTO:flat_walking_1} and Fig.\ref{fig:ARTO:flat_walking_2} illustrate the CoM position and active support foot against time in 3D during the two locomotion tasks: walking forward and sideways, respectively. The results show that in both scenarios the CoM remains approximately at constant height and follows a serpentine pattern on the transverse plane.

\subsection{Push Recovery on Flat Ground}
\begin{figure}[!htb]
  \centering
  \subfloat[]{
  \includegraphics[width=0.45\linewidth]{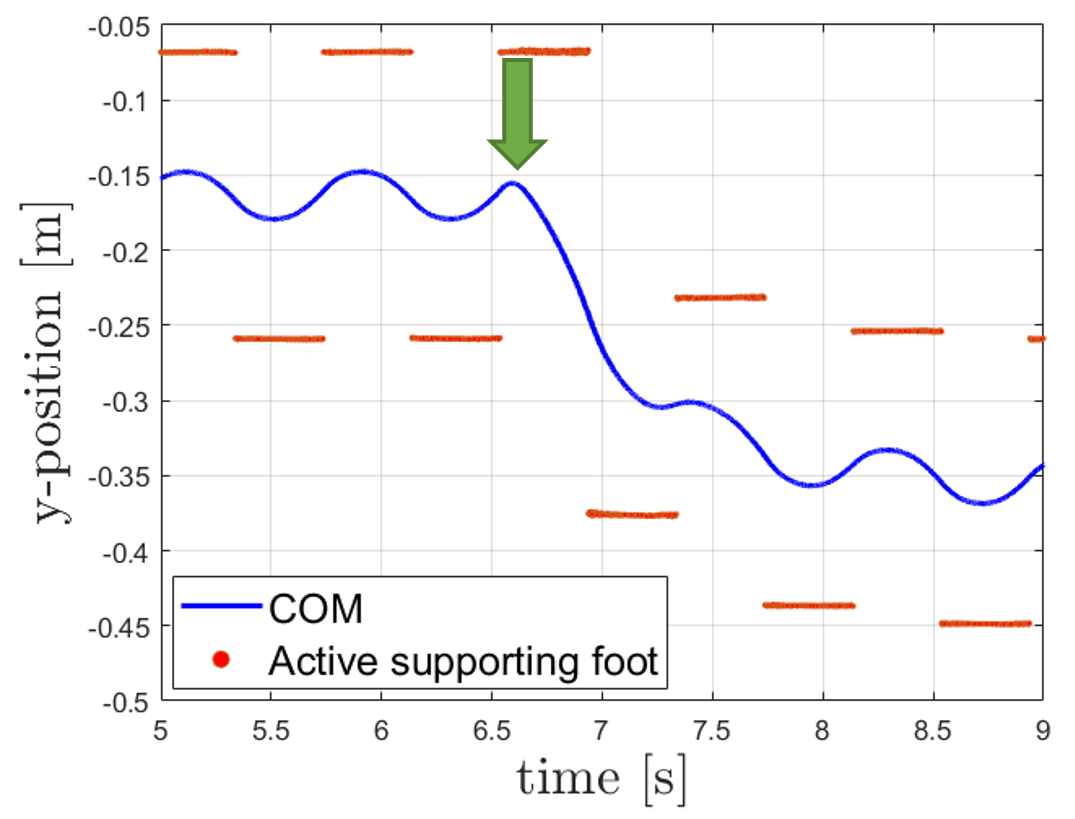}}
\subfloat[]{
  \includegraphics[width=0.45\linewidth]{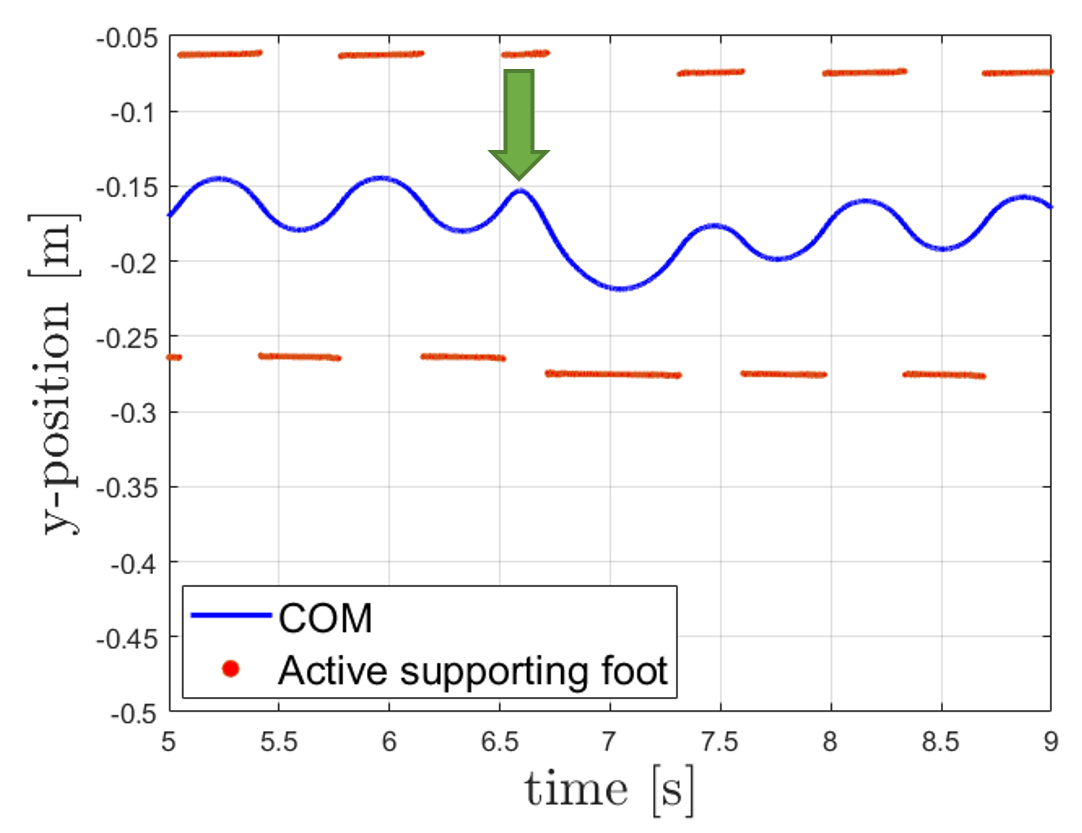}}
  \caption{Push recovery comparison between No-Time-Adp (a) and ARTO-AL (b). The figure highlights the behaviour between a planner that does not optimize time (a) and one that optimizes both footstep position and duration (b). The red arrow represents a 30 N side push of duration 0.1 s.}
  \label{fig:ARTO:flat_push_1}
\end{figure}

\begin{figure}[!htb]
    \centering
    \includegraphics[scale=0.95, trim={8.5cm 8cm 8.5cm 8cm}, clip]{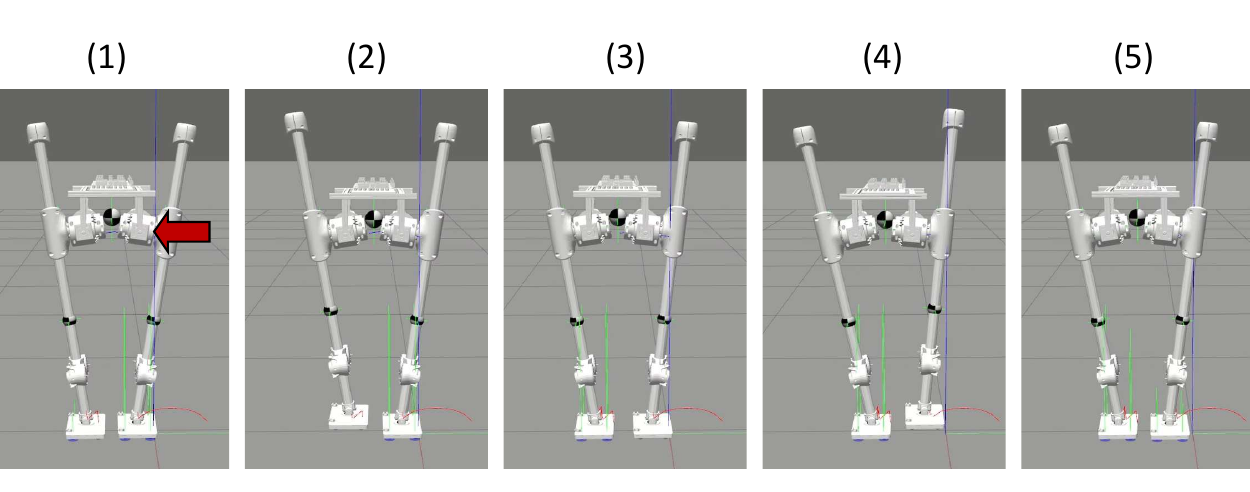}
    \caption{Push recovery using ARTO-AL. The model of SLIDER is stepping in place and is pushed sideways with a force of 30 N on the base and with a duration of 0.1 s. Red arrow indicates the direction of push. (1-3-5) show the moments corresponding to the switch of active supporting foot. (2-4) correspond to the sideways steps of the robot’s CoM to absorb the external force.}
    \label{fig:ARTO:flat_push_2}
\end{figure}

The push recovery experiment consisted of the analysis of SLIDER’s performances to external forces to compare two footstep planners: Footstep planner with no time adaptation (No-Time-Adp), and ARTO-AL. The aim of the experiment was to observe the push recovery of one footstep planner that adapts time and one that does not, demonstrating that optimizing both position and time offers better performances to external disturbances than when time is not optimized. For the first part of the experiment,SLIDER was stepping in place and then pushed laterally with a force of 30 N. Fig.\ref{fig:ARTO:flat_push_1} shows the comparison between a method that does not optimize time and one that does, i.e. No-Time-Adp and ARTO-AL, respectively.
Fig.\ref{fig:ARTO:flat_push_2} illustrates the five recovery stages performed by SLIDER using ARTO-AL with a 30 N push. Compared to No-Time-Adp, ARTO-AL is able to stabilise the robot after the push in 2 steps, returning to a normal stepping in place behaviour. However, without time adaptation, the robot requires at least 5 steps to return to normal stepping in place.

In the second experiment, the push recovery analysis was performed with the five footstep planners by measuring the maximum push force before failing. The five planners are: No-Time-Adp, IPOPT only, AL-only, DCM and ARTO-AL. The DCM step planner is adapted from the open source code \footnote{\url{https://github.com/machines-in-motion/reactive_planners}} which is the implementation of the paper \cite{Khadiv2016}. The robot was pushed in 8 directions spaced 45 degrees from each other. As a force is applied, the robot needs to react by taking a step in the same direction of the push, in order to prevent its body from falling. A successful push recovery consists of the robot returning to a stable walk after the push. The forces were applied at the touchdown moment of the left foot in a direction from left to right, using an impulse with a duration of 0.1 s, as shown in Fig.\ref{fig:ARTO:flat_push_2}. Overall, the maximum resist force is bigger in angle range $[0^{\circ}, 180^{\circ})$ than angle range of $[180^{\circ}, 360^{\circ})$ because of the direction of push force. ARTO-AL has the best performance over all planners.

\begin{figure}[!htb]
    \centering
    \includegraphics[scale=0.45]{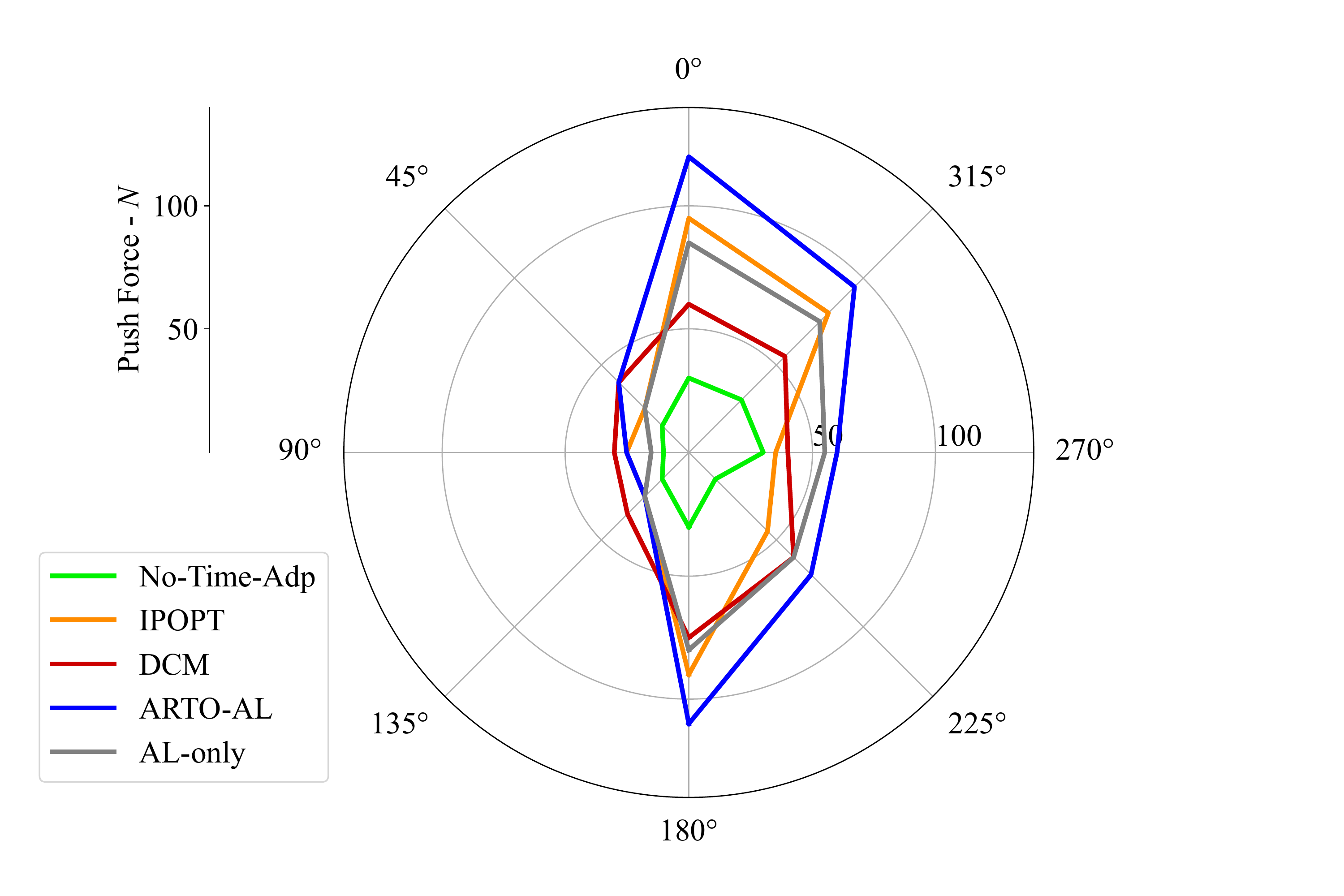}
    \caption{Maximum push recovery on flat ground with forces parallel to the transverse plane. The impulse was applied with a duration of 0.1 s on the base while SLIDER was stepping in place. 0$^{\circ}$ corresponds to the robot being pushed forward.}
    \label{fig:ARTO:flat_push_3}
\end{figure}

\subsection{Push Recovery on Uneven Terrain}
\begin{figure}[!htb]
    \centering
    \includegraphics[scale=0.45]{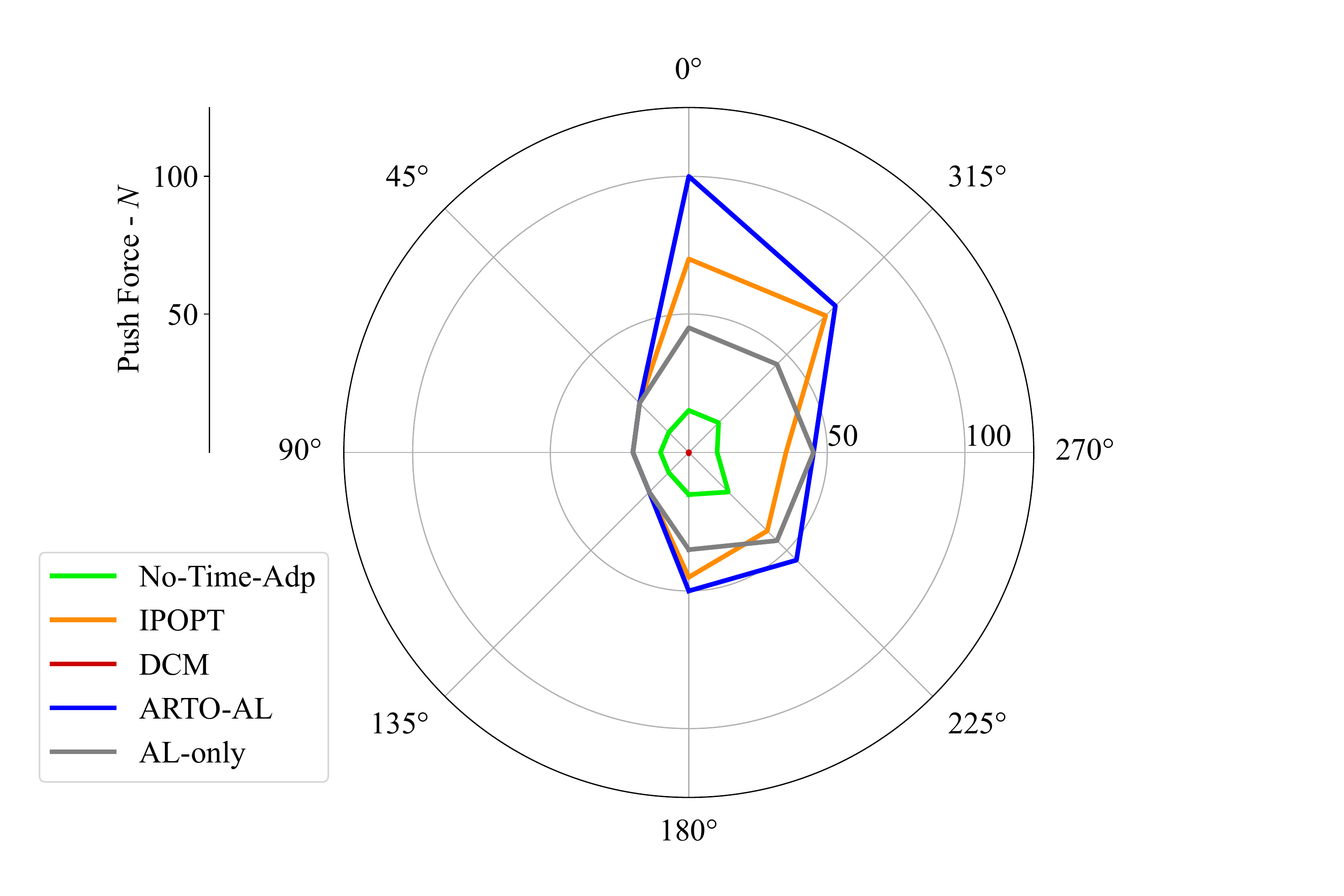}
    \caption{Maximum push recovery on inclined ground with the force vector parallel to the transverse plane. The impulse of the disturbance lasted 0.1 s and was applied while walking on a 10$^{\circ}$ ramp at 0.1 m/s. 0$^{\circ}$ corresponds to the robot being pushed forward.}
    \label{fig:ARTO:uneven_push_1}
\end{figure}

\begin{figure}[!htb]
    \centering
    \includegraphics[scale=0.35]{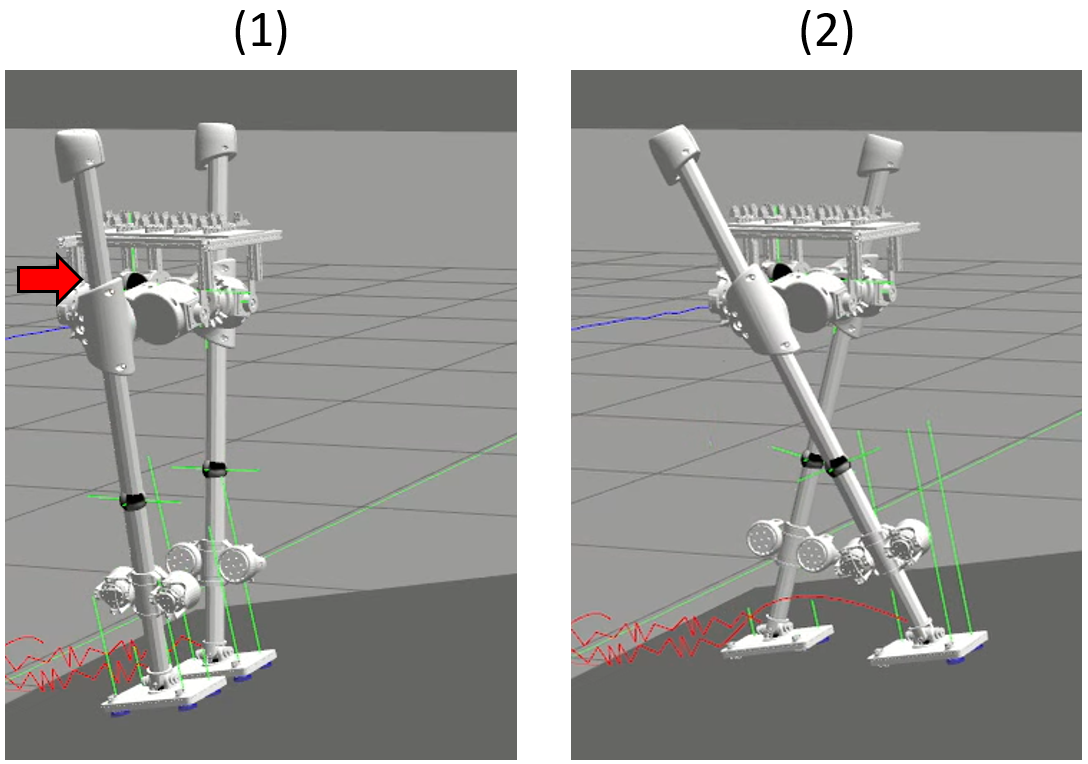}
    \caption{Reaction step to a 70 N push when walking on a 10$^{\circ}$ slope with ARTO-AL. Red arrow indicates the push direction.}
    \label{fig:ARTO:uneven_push_2}
\end{figure}

Fig.\ref{fig:ARTO:uneven_push_1} shows the push recovery performance of 5 planners while SLIDER walks up a 10$^{\circ}$ ramp given the slope angle. The maximum force of forward push is bigger than backward push. The reason is that the foot cannot touch the ground as a result of leg length limit when pushed backward. ARTO-AL performs best among 5 planners. Fig.\ref{fig:ARTO:uneven_push_2} shows the reaction of SLIDER when pushed forward with ARTO-AL. The DCM planner is not compared because constructing a plane needs two steps prediction ahead. 

\subsection{Terrain-Aware Walking}
\begin{figure}[!htb]
  \centering
  \subfloat[]{
  \includegraphics[width=0.475\linewidth]{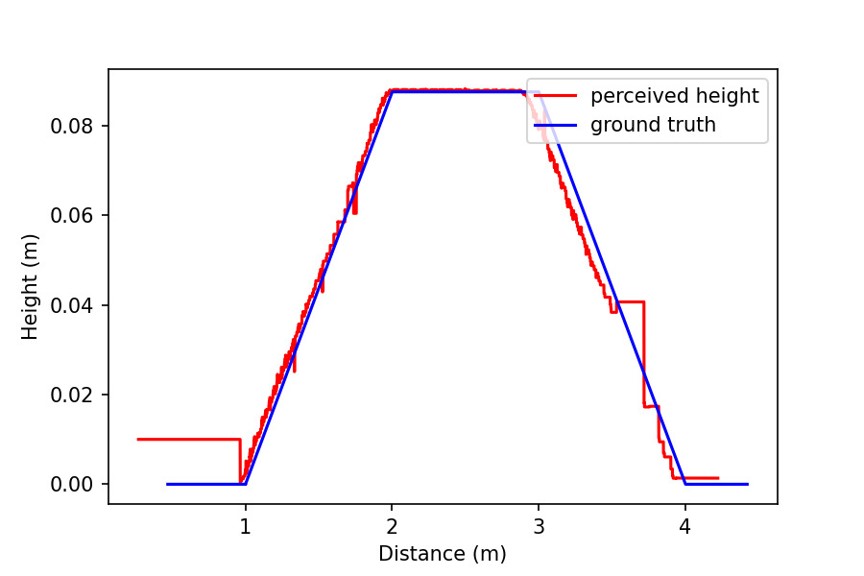}}
\subfloat[]{
  \includegraphics[width=0.475\linewidth]{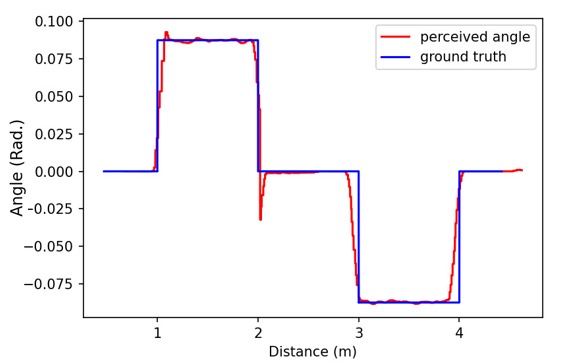}}
  \caption{The estimated height and slope angle from height map compared with the ground truth, when the robot walks up and down on a 5$^\circ$ slope. (a) The estimated height vs ground truth. (b) The estimated slope angle vs ground truth.}
  \label{fig:ARTO:slope_vision}
\end{figure}

Fig.\ref{fig:ARTO:slope_vision} shows the comparison between the estimated height and slope angle from the height map compared with the ground truth. Because early contact or late contact can easily make the robot unstable, dynamic gains is utilized: the gains are designed to interpolate linearly for a smooth switch between stance and swing foot during a short time period. The average error is 0.0086~m for height and 1.09$^{\circ}$ for the slope angle. Fig. \ref{fig:ARTO:slope_vision} shows the success rate of walking with three different conditions. Benchmark: no perception and no dynamic gains. ARTO-AL+No DG: ARTO-AL without dynamic gains. ARTO-AL + DG: ARTO-AL with dynamic gains. As can be seen the inclusion of terrain-aware ARTO-AL provides large improvements in performance over the benchmarks bringing the critical failure rate for steps from 3 to 4~cm. The largest improvements can be seen on slopes where the critical failure rate is above 10$^{\circ}$ where it is 2.5$^{\circ}$ for benchmark. Dynamic gains further improves the performance, for step height of 3~cm the success rate reaches around 90$\%$. The success rate also increases to 90$\%$ with dynamic gains when the robot walks on the 10$^{\circ}$ slope.
\begin{figure}[!htb]
  \centering
  \subfloat[]{
  \includegraphics[width=0.5\linewidth]{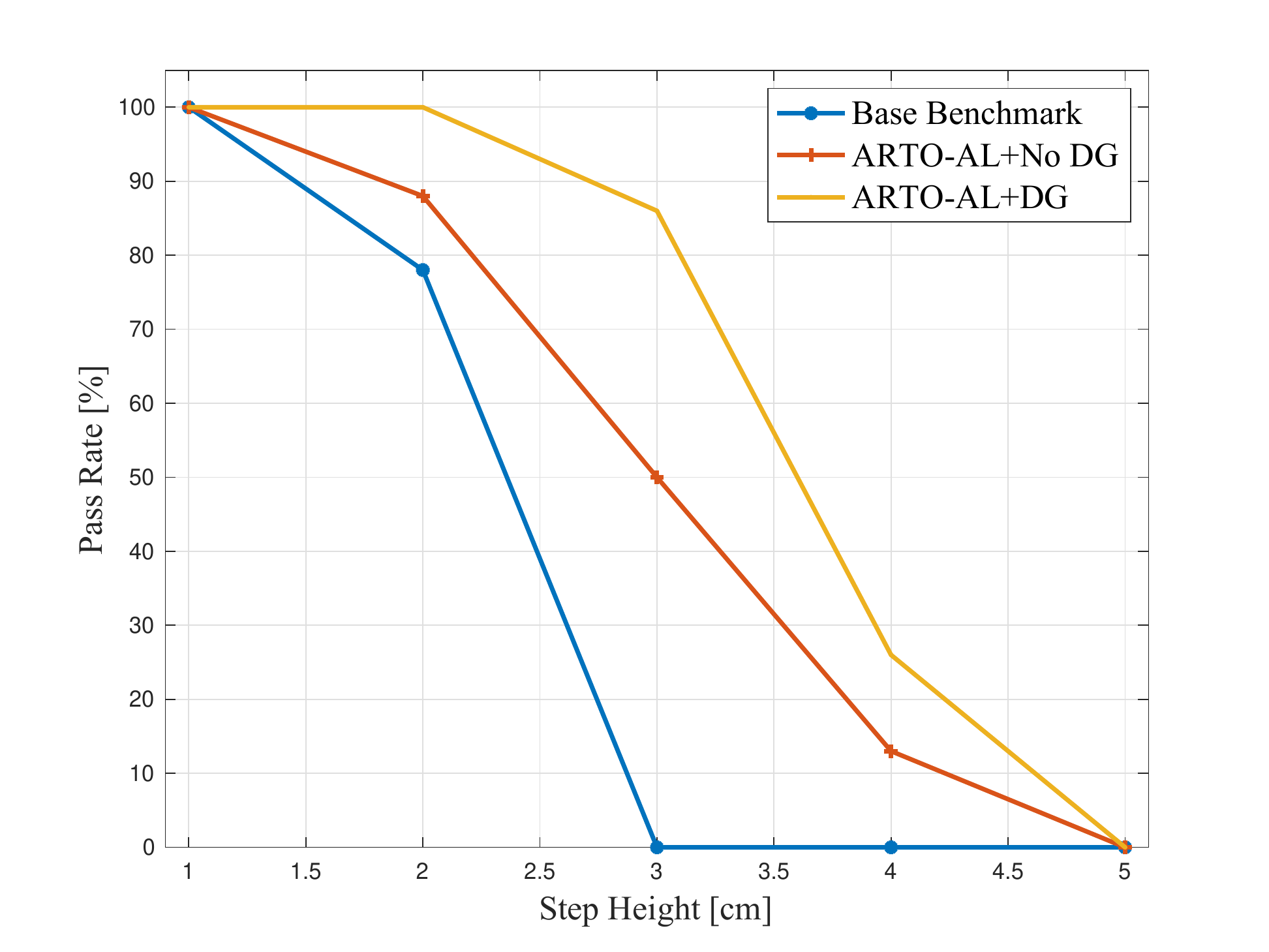}}
\subfloat[]{
  \includegraphics[width=0.5\linewidth]{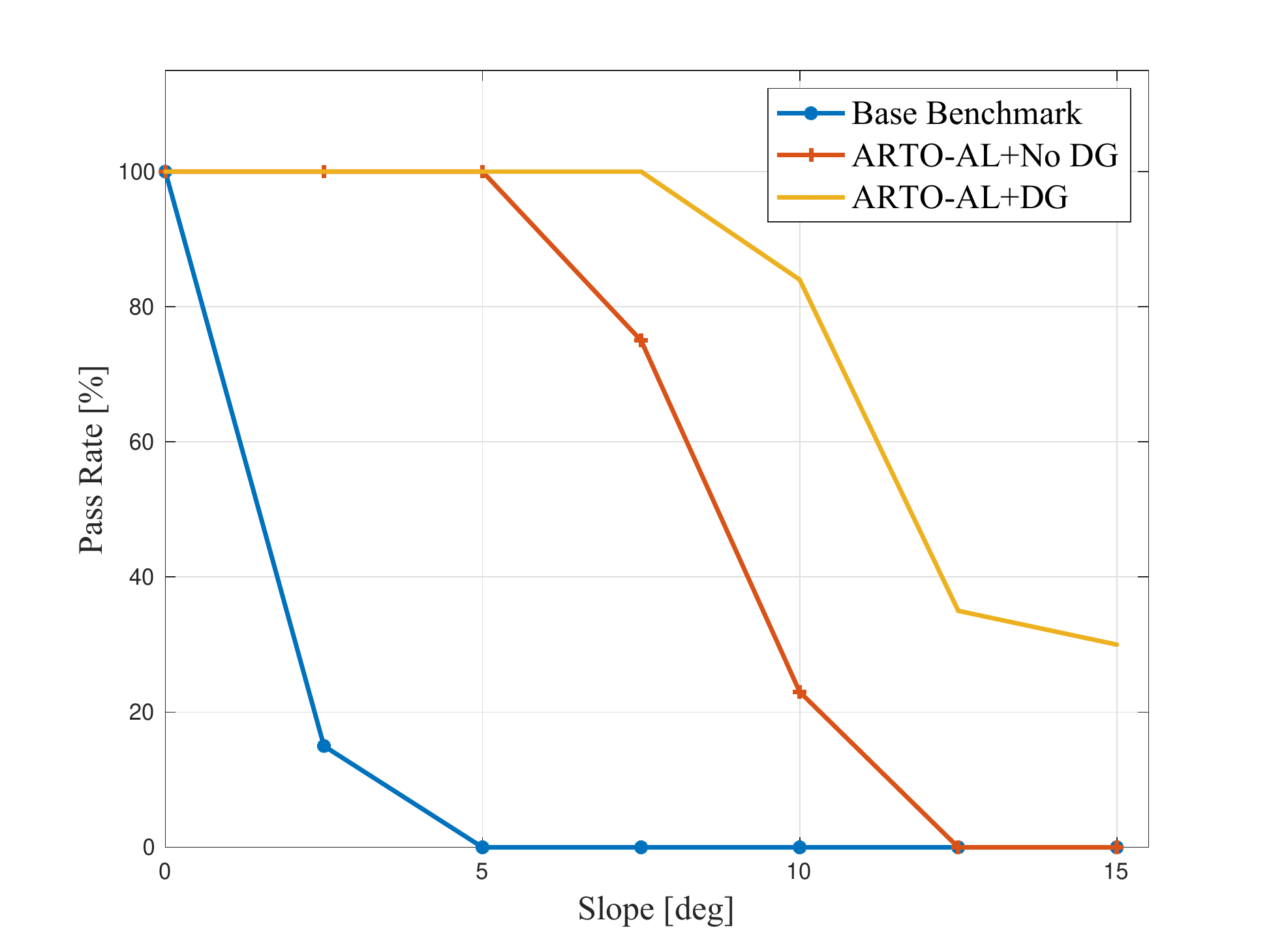}}
  \caption{Plots of success rate when walking in three conditions. Base Benchmark: no perception and no dynamic gains. ARTO-AL+No DG: ARTO-AL without dynamic gains. ARTO-AL + DG: ARTO-AL with dynamic gains. (a), Success rate with different step height. (b), Success rate with different slope angles.}
  \label{fig:ARTO:vision_compare}
\end{figure}
\section{DISCUSSION}
    \subsection{DCM vs full LIP dynamics} In \cite{Khadiv2020} the footstep location and timing adaptation problem is formulated using DCM and it can be solved with QP. While solving with QP gives the real-time guarantee, it is only viable for \textit{one} step prediction. Because of this, the approach using DCM cannot easily extend to uneven terrain walking. In the meantime, as shown in the push recovery result, approach using DCM performs worse than ARTO-AL in most cases. Furthermore, using DCM needs extra steps and heuristics to compute DCM offset. While our approach is straightforward and requires no extra steps.
    \subsection{Using IPOPT as initial guess} Using IPOPT as a initial guess to bootstrap the optimization solver has greatly improved the performance. On average, ARTO-AL performs 28\% better than IPOPT and AL-only, both on flat ground and slopes. IPOPT and AL-only perform similarly, IPOPT performs slightly better on forward and backward push while AL-only performs slightly better on horizontal pushes. ARTO-AL which is the combination of the two approaches performs better in all cases.
    \subsection{Uneven terrain walking} We showed theoretically and experimentally that our approach can be extended to 3D and therefore applied to uneven terrain walking, under the assumption that the local terrain around the robot can be described as a plane. This assumption is valid for terrains where gradient changes are not sharp. Furthermore, our approach does not constrain footsteps in specific regions so this approach may fail in scenarios like stepping stones \cite{nguyen2017dynamic}.

\section{CONCLUSION}
This paper proposes a terrain-aware footstep planning algorithm that can adapts step position and timing based on the LIP dynamics. With an asynchronous structure which combines an augmented lagrangian solver with analytical gradient descent and IPOPT optimizer, the planner can achieves real-time performance with 200~Hz update frequency. Furthermore, we extend the planner from flat ground walking to uneven terrain with perceptive information. We have shown experimental results of the robot walking on flat ground, uneven terrains and push recovery on both flat ground and a slope with 10$^{\circ}$, demonstrating its performance and robustness to disturbances.
Future work will include realising real-time footstep location and timing optimization with more complex models such as centroidal models. Another interesting will extend our work in more constrained environments which may contain stepping stones and obstacles.
\section*{APPENDIX}
\subsubsection{Analytical gradients of footstep position}
\begin{multline*}
    \frac{\partial x_j}{\partial u_{x, k}} =
    \begin{cases}
        0 & j \leq k\\
        -\frac{1}{2}(e^{\omega \Delta t_{j-1}} + e^{-\omega \Delta t_{j-1}}) + 1 & j = k + 1\\\\
        \frac{1}{2}\frac{\partial x_{j-1}}{\partial u_{x, k}}(e^{\omega \Delta t_{j-1}} + e^{-\omega \Delta t_{j-1}}) +& j \geq k+2\\
        \frac{1}{2\omega}\frac{\partial \dot{x}_{j-1}}{\partial u_{x, k}}(e^{\omega \Delta t_{j-1}} - e^{-\omega \Delta t_{j-1}})
    \end{cases}
\end{multline*}

\begin{multline*}
    \frac{\partial \dot{x}_j}{\partial u_{x, k}} =
    \begin{cases}
        0 & j \leq k\\
        -\frac{1}{2}\omega(e^{\omega \Delta t_{j-1}} - e^{-\omega \Delta t_{j-1}}) & j = k + 1\\\\
        \frac{\omega}{2}\frac{\partial x_{j-1}}{\partial u_{x, k}}(e^{\omega \Delta t_{j-1}} - e^{-\omega \Delta t_{j-1}}) + & j \geq k+2
        \\\frac{1}{2}\frac{\partial \dot{x}_{j-1}}{\partial u_{x, k}}(e^{\omega \Delta t_{j-1}} + e^{-\omega \Delta t_{j-1}})
    \end{cases}
\end{multline*}
With identical formulation in the $y$ direction.
\subsubsection{Analytical gradients of footstep duration}
\begin{multline*}
    \frac{\partial x_j}{\partial \Delta t_k} =
    \begin{cases}
        0 & j < k
        \\
        \dot{x}_j & j=k
        \\
        \frac{1}{2}\frac{\partial x_{j-1}}{\partial \Delta t_{k}}(e^{\omega \Delta t_j} + e^{-\omega \Delta t_j}) + & j \geq k+1
        \\
        \frac{1}{2 \omega}\frac{\partial \dot{x}_{j-1}}{\partial \Delta t_{k}}(e^{\omega \Delta t_j} - e^{-\omega \Delta t_j})
    \end{cases}
\end{multline*}\vspace{-0.5cm}
\begin{multline*}
    \frac{\partial \dot{x}_j}{\partial \Delta t_k} =
    \begin{cases}
        0 & j < k
        \\
        \ddot{x}_j & j=k
        \\
        \frac{\omega}{2}\frac{\partial x_{j-1}}{\partial \Delta t_{k}}(e^{\omega \Delta t_j} - e^{-\omega \Delta t_j}) + & j \geq k+1
        \\
        \frac{1}{2}\frac{\partial \dot{x}_{j-1}}{\partial \Delta t_{k}}(e^{\omega \Delta t_j} + e^{-\omega \Delta t_j})
    \end{cases}
\end{multline*}
Again, with identical formulation in the $y$ direction.
\section*{ACKNOWLEDGEMENT}
\bibliography{mybibfile}

\end{document}